\documentclass[11pt]{article}

\usepackage[preprint]{acl}

\usepackage{times}
\usepackage{latexsym}

\usepackage[T1]{fontenc}

\usepackage[utf8]{inputenc}

\usepackage{microtype}

\usepackage{inconsolata}

\usepackage{graphicx}

\usepackage{xcolor}

\edef\aclrm{\rmdefault}\edef\aclsf{\sfdefault}\edef\acltt{\ttdefault}
\usepackage{fontspec}

\renewcommand{\rmdefault}{\aclrm}\renewcommand{\sfdefault}{\aclsf}\renewcommand{\ttdefault}{\acltt}
\newfontfamily\bengalifont[Path=./,Script=Bengali,UprightFont=NotoSerifBengali-Regular.ttf]{NotoSerifBengali}
\newcommand{\bn}[1]{{\bengalifont #1}}

\usepackage{float}      
\usepackage{placeins}   %
\usepackage{adjustbox} 
\usepackage{array}
\usepackage{booktabs}        
\usepackage{colortbl}

\definecolor{vec}{HTML}{0F766E}
\definecolor{vecbg}{HTML}{E6F6F4}
\definecolor{full}{HTML}{C2410C}
\definecolor{fullbg}{HTML}{FFF1E8}
\newcommand{\vecv}[1]{\colorbox{vecbg}{\textcolor{vec}{#1}}}
\newcommand{\fullv}[1]{\colorbox{fullbg}{\textcolor{full}{#1}}}

\title{Pole or Vector? Encoders Capture Bangla Compound Verbs, but LLMs Read Them Literally}

\author{
\textbf{Md. Asaduzzaman Shuvo\textsuperscript{1},
Md Mehrab Karim Antor\textsuperscript{1},
Md. Adnanul Islam\textsuperscript{1},}\\
\textbf{Shahanur Rahman Bappy\textsuperscript{1},
Sajid Ahamed\textsuperscript{1},
Mahfuz Hasan Reza\textsuperscript{1},
Md. Delwar Shahadat Deepu\textsuperscript{2}}
\\[4pt]
\textsuperscript{1}United International University, Dhaka, Bangladesh \\
\textsuperscript{2}Northern University Bangladesh, Dhaka, Bangladesh
\\[4pt]
{\small \textbf{Emails:} \texttt{\{ashuvo221104, mantor221595, mislam221096\}@bscse.uiu.ac.bd},} \\
{\small \texttt{\{sbappy211002, sahamed221252, mreza221203\}@bscse.uiu.ac.bd},} \\
{\small \texttt{deepu@nub.ac.bd}}
}

\begin{document}
\maketitle
\begin{abstract}
In Bangla, a main verb is often followed by one of a small set of vector
verbs. These verbs lose their own meaning and add a shade of meaning instead,
such as completion. For example, \textit{likhe phello} means that someone
finished writing, not that they wrote and threw something. The same two verbs
can also describe two separate events, so only the context shows which meaning
is meant. This makes compound verbs a hard test of how language models handle
multiword expressions. We present \textsc{BanglaCV-Probe}, a benchmark of 17
vector verbs with 1,044 minimal pairs, 204 entailment items and 197 pairs of
correct and incorrect combinations, all checked by native speakers. We test 26
models (encoders, LLMs and translation systems) on four skills: telling the two
meanings apart internally, understanding what a vector adds, preferring correct
combinations and keeping the vector's meaning in translation. Probes on five
encoders tell the two meanings apart and still reach up to 93\% accuracy when
trained on pairs without surface cues. Prompted models rarely use this
knowledge. Entailment accuracy stays between 59\% and 81\%. Several LLMs answer
Yes or No to almost every item, and nine of ten models miss that someone who
ate up the rice left none. Most instruction-tuned LLMs translate vectors
literally in 15--20\% of vector sentences and in up to 55\% for \textit{mora}
(die). On correct combinations, a small Bangla-only model (88\%) matches
BLOOM-7B1 (87\%) and both beat Qwen2.5-7B (74\%).
\end{abstract}

\section{Introduction}

\begin{figure}[t]
  \centering
  \includegraphics[width=\columnwidth]{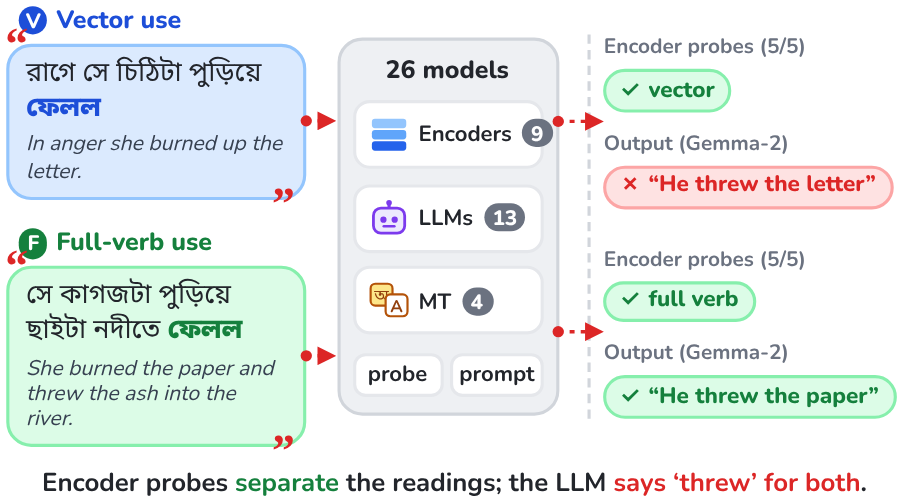}
  \caption{The same V2 \bn{ফেলা} (throw) as a vector (\textsc{v}) and as a
  full verb (\textsc{f}). Probes on the V2's hidden state get both sentences
  right in all five encoders. The LLM Gemma-2-2B-IT still translates both with
  \textit{threw}. Translations are real model outputs.}
  \label{fig:teaser}
\end{figure}

Multiword expressions (MWEs) are word groups whose meaning is not just the sum
of their parts. They have long been \textit{a pain in the neck} for NLP
\citep{SagBBCF02,constant-etal-2017-survey}. Large language models handle them
better than older systems but still unevenly, and often follow surface
patterns instead of meaning \citep{miletic-walde-2024-semantics}. Most of this
evidence comes from English. For Bangla, even idioms are still hard: no
current LLM goes above 50\% accuracy, while humans reach 83.4\%
\citep{sakhawat-etal-2026-words}.

A common Bangla MWE, also found in other South Asian languages, is the
compound verb \citep{hook1974compound,abbi1991semantics}. A main verb in
conjunctive form (the pole, V1) is followed by one of a small set of vector
verbs (V2). The vector loses its own meaning and adds a shade of meaning
instead, such as completion \citep{butt2010light,paul2003composition}. In
\bn{লিখে ফেলেছে} \textit{likhe pheleche} (has finished writing) nothing is
thrown, although \bn{ফেলা} alone means throw. But the same two verbs can also
keep their own meanings:

\begin{center}
\small
\setlength{\fboxsep}{1pt}
\begin{tabular}{@{}l@{\hspace{4pt}}p{0.86\columnwidth}@{}}
(1) & \bn{সে চিঠিটা \vecv{লিখে দিল}।} \hfill \textcolor{vec}{\textsc{vector}}\\
    & \textit{She wrote the letter (for someone).}\\[3pt]
(2) & \bn{সে চিঠিটা \fullv{লিখে} আমার হাতে \fullv{দিল}।} \hfill \textcolor{full}{\textsc{full verb}}\\
    & \textit{She wrote the letter and handed it to me.}
\end{tabular}
\end{center}

\noindent The context must show which use is meant
\citep{mukhopadhayay-etal-2012-automatic}, so a model cannot just remember
word strings. When it gets this wrong, the error is easy to see. One model we
test translates \bn{রাগে সে চিঠিটা পুড়িয়ে ফেলল} (in anger she burned up the
letter) as \textit{he threw the letter in anger} (Figure~\ref{fig:teaser}).
Past work on Bangla compound verbs tried to find them in text
\citep{mukhopadhayay-etal-2012-automatic}, not to test what neural models
understand. Recent probing of light verbs covers only noun--verb pairs in
European languages \citep{franzon-etal-2026-light}.

We introduce \textsc{BanglaCV-Probe}, which covers the 17 vector verbs listed by
\citep{paul2003composition} and \citep{ghosh-2014-making}
(Appendix~\ref{app:vectors}) and has three parts (Table~\ref{tab:data}). As in earlier minimal-pair work
\citep{warstadt-etal-2020-blimp-benchmark}, we match each compound-verb
sentence with a full-verb sentence, and we add entailment items and pairs of
correct and incorrect combinations. We ask whether models (1)~tell the two
uses apart internally, (2)~understand what a vector adds, (3)~prefer correct
pole--vector combinations and (4)~keep the vector's meaning in translation.
We probe encoders for (1) and prompt or score all models for (2)--(4). We do
not trust probe accuracy alone \citep{hewitt-liang-2019-designing}, so we
compare it with random-weight models, a surface cue and hard pairs without
that cue.

The answers are very different. Encoders capture the difference: probes
trained on cue-free pairs reach up to 93\% accuracy, while random-weight
models stay near chance. Prompted models rarely use it. Entailment accuracy
stays between 59\% and 81\%, and nine of ten models miss that someone who ate
up all the rice left none. Most instruction-tuned LLMs translate vectors
literally in 15--20\% of vector sentences. For correct combinations, a small
Bangla-only model (88\%) matches BLOOM-7B1 (87\%) and both beat Qwen2.5-7B
(74\%).

This paper makes three contributions. (i)~We release \textsc{BanglaCV-Probe},
a benchmark of Bangla compound verbs with minimal pairs, entailment items and
combination pairs from written and real text. (ii)~We test 26 models
(encoders, LLMs and MT systems) on four skills, from inner representations to
translation, with checks for surface cues and answer bias. (iii)~We show that
encoders capture the difference but prompted models rarely use it, and that
some of the same vectors are read literally across tasks.
\section{Related Work}

\paragraph{Multiword expressions and light verbs in language models.}
MWEs have a long history in NLP \citep{constant-etal-2017-survey}, and recent
surveys find that transformer models capture MWE meaning unevenly
\citep{miletic-walde-2024-semantics}. In translation, models often translate
idioms word by word \citep{dankers-etal-2022-transformer}. Minimal pairs, a
common test of grammar knowledge
\citep{marvin-linzen-2018-targeted,warstadt-etal-2020-blimp-benchmark}, are now
also used for fixed phrases: \citep{franzon-etal-2026-light} show that models
tell light from full uses of the same verb (\textit{make a decision} vs.\
\textit{make a cake}) in English, Spanish and French. For Hindi,
\citep{jain-vaidya-2025-benchmark} find that causal models do poorly on
minimal pairs for verb--argument changes, and \citep{choudhary-etal-2026-llms}
show that LLMs cannot reliably predict whether speakers use a Hindi verb or an
English verb with the light verb \textit{karna}. None of this work covers
verb--verb compounds, where the light part is itself a verb that has lost its
meaning, or Bangla.

\paragraph{Bangla compound verbs.}
Linguistic work describes Bangla compound verbs in detail.
\citep{paul2003composition} lists the vector verbs and argues that a V1 joins
only with vectors whose meaning fits. \citep{ghosh-2014-making} builds verb
frames for 15 vectors, and \citep{bhattacharja2018formation} argues that
complex predicates are formed in the lexicon. Computational work has mostly
tried to find compound verbs in text, with hand-written rules
\citep{das-etal-2010-automatic,mukhopadhayay-etal-2012-automatic} or
IndoWordNet features \citep{singh-etal-2016-detection}. The closest to our
questions is \citep{dasgupta-etal-2015-compositionality}, who link the meaning
of compound verbs to how people process them, using a count-based vector
model. None of this work uses contextual language models, and to our knowledge
no labelled benchmark of Bangla compound verbs exists.

\paragraph{Testing Bangla language models beyond fluent output.}
More and more benchmarks test Bangla meaning that the surface form does not
show. Idiom understanding is still far below human level
\citep{sakhawat-etal-2026-words}. Telling grammatical word doubling from
repeated words in speech needs context, and fine-tuned encoders do better at
it than few-shot LLMs \citep{arpa2025distinguishing}.
\citep{shuvo-islam-2026-name} find a dominant-meaning bias for words that are
both names and common nouns: models choose the noun reading unless asked to
compare the two. \citep{shuvo-etal-2026-polite} show that LLMs also use Bangla
honorifics inconsistently. Compound verbs raise a similar problem, since the
same verb has a literal and a grammatical reading. We find the same bias
towards the literal reading in model outputs, even though probes show that
encoders capture the grammatical reading.

\section{BanglaCV-Probe}
\label{sec:data}

\textsc{BanglaCV-Probe} covers all 17 vector verbs listed by
\citep{paul2003composition, ghosh-2014-making} (Appendix~\ref{app:vectors}) and has three parts,
one for each kind of question (Table~\ref{tab:data}).
\paragraph{Minimal pairs.}
Each pair matches a vector sentence with a sentence in which the same V2 is a full verb: \bn{ছেলেটা পুরো ভাতটা খেয়ে \textbf{ফেলল}} (the boy ate up all the rice) vs.\ \bn{ছেলেটা কলাটা খেয়ে খোসাটা রাস্তায় \textbf{ফেলল}} (the boy ate the
banana and threw the peel on the road).

\definecolor{rqA}{HTML}{0F766E}   
\definecolor{rqB}{HTML}{6D28D9}   
\definecolor{rqC}{HTML}{B45309}   
\definecolor{rqD}{HTML}{C2410C}   
\newcommand{\rqtag}[2]{{\setlength{\fboxsep}{1.2pt}\colorbox{#1!13}{\textcolor{#1}{\scriptsize\textbf{#2}}}}}

\begin{table}[t]
\centering
\small
\setlength{\tabcolsep}{4pt}
\begin{tabular}{@{}lrl@{}}
\toprule
Part & Items & Used for \\
\midrule
\rowcolor{rqA!6}
\textbf{Minimal pairs} {\footnotesize(vector vs.\ full verb)} & 1,044 & \rqtag{rqA}{RQ1} \rqtag{rqD}{RQ4} \\
\quad {\footnotesize adjacent / hard (cue-free)}  & {\footnotesize 74}  & \\
\quad {\footnotesize from real text}              & {\footnotesize 103} & \\
\rowcolor{rqB!6}
\textbf{Entailment items}                          & 204 & \rqtag{rqB}{RQ2} \\
\rowcolor{rqC!6}
\textbf{Selectional pairs} {\footnotesize(valid vs.\ invalid)} & 197 & \rqtag{rqC}{RQ3} \\
\midrule
Total                                              & 1,445 & \\
\bottomrule
\end{tabular}
\caption{\textsc{BanglaCV-Probe}. All parts cover the 17 vector verbs
(64--95 items per verb). Colours match the research questions in
Table~\ref{tab:models}.}
\label{tab:data}
\end{table}

In most pairs, the word before V2 gives the answer away, because it is a
conjunctive participle only in the vector sentence. So we added 74
adjacent pairs in which V1 comes right before V2 in both sentences,
with V2 in the same position, so that the wider context must decide. Of the
1,044 pairs, 941 were written by us and 103 come from real text. For the real
text, we chose and paired sentences by hand from 260k sentences of
public-domain fiction and news. The pairs contain 552 different V1s.

\paragraph{Entailment items.}
We wrote 204 premise--hypothesis pairs that test what a vector adds. The
hypothesis either states the vector's meaning (e.g.\ nothing is left after
\textit{ate up}) or states the V2's literal meaning, which a vector
premise does not entail. For example, \bn{শিশুটা হঠাৎ কেঁদে উঠল} (the child
suddenly burst into tears) does not mean that the child climbed anywhere. We
call these 64 items literal traps. Full-verb and simple-verb premises
serve as controls.

\paragraph{Selectional pairs.}
Vectors do not combine freely. For example, \bn{ফেলা}, which marks
completion, does not fit with a phrase of duration
\citep{paul2003composition}. Each of the 197 pairs puts a correct sentence
next to an incorrect one that differs in one word or phrase, such as
\bn{এক ঘণ্টার \textbf{মধ্যে} লিখে ফেলল} (wrote it within an hour) vs.\
*\bn{এক ঘণ্টা \textbf{ধরে} লিখে ফেলল} (for an hour). The pairs cover seven
rule types, such as duration, doing something for someone else and whether V1
needs a doer (Appendix~\ref{app:data}).

\paragraph{Validation and splits.}
Three native speakers checked every item on their own (Fleiss'
$\kappa = 0.92$). For probing, we split the written non-adjacent pairs into
train, development and test sets (609 / 82 / 176), so that no V1 appears in
more than one set. The adjacent and corpus pairs are used only for testing.

\section{Experimental Setup}
\label{sec:setup}

We test 26 public models (Table~\ref{tab:models}; versions and references in
Appendix~\ref{app:models}). We do not fine-tune any
model on our data; only the probes in RQ1 are trained. Large models run on two
T4 GPUs in 4-bit NF4 precision \citep{dettmers2023qlora}.

\definecolor{rqA}{HTML}{0F766E}   
\definecolor{rqB}{HTML}{6D28D9}   
\definecolor{rqC}{HTML}{B45309}   
\definecolor{rqD}{HTML}{C2410C}   
\newcommand{\ya}{\textcolor{rqA}{\large$\bullet$}}
\newcommand{\yb}{\textcolor{rqB}{\large$\bullet$}}
\newcommand{\yc}{\textcolor{rqC}{\large$\bullet$}}
\newcommand{\yd}{\textcolor{rqD}{\large$\bullet$}}
\definecolor{encBg}{HTML}{E8EEF7} \definecolor{encFg}{HTML}{1E3A8A}   
\definecolor{nliBg}{HTML}{F5E9F2} \definecolor{nliFg}{HTML}{86198F}   
\definecolor{llmBg}{HTML}{E9F5EC} \definecolor{llmFg}{HTML}{166534}   
\definecolor{mtBg}{HTML}{F3F1EC}  \definecolor{mtFg}{HTML}{57534E}    
\newcommand{\famrow}[3]{\rowcolor{#1}\multicolumn{5}{@{\hspace{2pt}}l}{\textcolor{#2}{\textbf{\textit{#3}}}}}

\begin{table}[t]
\centering
\footnotesize
\setlength{\tabcolsep}{4pt}
\begin{tabular}{@{}lcccc@{}}
\toprule
 & \textbf{RQ1} & \textbf{RQ2} & \textbf{RQ3} & \textbf{RQ4} \\[-1pt]
 & \scriptsize probe & \scriptsize NLI & \scriptsize surpr. & \scriptsize MT \\
\midrule
\famrow{encBg}{encFg}{Encoders} \\
\quad BanglaBERT, mBERT, MuRIL      & \ya & & \yc$^\dagger$ & \\
\quad XLM-R base, large             & \ya & & \yc & \\
\famrow{nliBg}{nliFg}{NLI classifiers} \\
\quad mDeBERTa ($\times$2), XLM-R-l & & \yb & & \\
\famrow{llmBg}{llmFg}{Decoder LLMs} \\
\quad Qwen2.5 (0.5--7B)             & & \yb$^\ast$ & \yc & \yd$^\ast$ \\
\quad BLOOM(Z) 7B, Gemma-2 2B       & & \yb$^\ast$ & \yc & \yd$^\ast$ \\
\quad Llama-3.1 8B                  & & \yb$^\ast$ & & \yd$^\ast$ \\
\famrow{mtBg}{mtFg}{MT systems} \\
\quad NLLB-200 (0.6--3.3B)          & & & & \yd \\
\quad BanglaT5-NMT                  & & & & \yd \\
\bottomrule
\end{tabular}
\caption{The 26 models and where they are used. $^\ast$Instruction-tuned
version (BLOOMZ for BLOOM); RQ4 uses only Qwen2.5-7B. $^\dagger$In RQ3,
BanglaBERT is replaced by its generator, which has a masked-language-model
head.}
\label{tab:models}
\end{table}
\paragraph{RQ1: Probing.}
For each sentence, we take the mean of the subword vectors of V2 at every layer
and train a logistic-regression probe to label it as a vector or a full verb.
We choose the layer and the regularisation strength on the development set.
A probe can succeed by using simple patterns
\citep{hewitt-liang-2019-designing}, so we compare it with three baselines:
\begin{itemize}
  \setlength\itemsep{0pt}
  \item the same model with random weights;
  \item a one-feature surface cue: is the word before V2 a conjunctive participle?
  \item the majority class.
\end{itemize}
We report accuracy on the test, hard and corpus sets. The hard pairs are a
different kind of pair from the training pairs, so we also run a 5-fold
cross-validation inside the hard pairs and choose the layer inside each fold.

\paragraph{RQ2: Entailment.}
The NLI classifiers predict entailment when it is their most likely
class. The instruction-tuned LLMs answer each item zero-shot. We compare the
probabilities of \textit{Yes} and \textit{No} as the first answer token,
average them over an English and a Bangla instruction, and predict
\textit{Yes} above 0.5. Many models prefer one answer whatever the item is.
So besides accuracy and macro-F1, we report AUC, which does not depend on a
cut-off, and the share of \textit{Yes} answers.

\paragraph{RQ3: Selectional preferences.}
A model is correct on a pair if it gives the correct sentence a higher
probability than the incorrect one. For causal models, we use the summed token
log-probability. For masked models, we use the pseudo-log-likelihood
\citep{salazar-etal-2020-masked} in the word-level version of
\citep{kauf-ivanova-2023-better}. Summed probabilities favour short sentences,
so we compare each model with a baseline that always picks the sentence with
fewer tokens.

\paragraph{RQ4: Translation.}
We translate both sentences of every minimal pair into English, with beam
search for the MT systems and greedy zero-shot decoding for the LLMs. To find
literal translations, we list the English words that express each V2's own
meaning (e.g.\ \textit{throw, drop} for \bn{ফেলা}). In our reference
translations, these words appear in 5\% of vector sentences and 91\% of
full-verb sentences, so they separate the two uses well. We report two rates:
\begin{itemize}
  \setlength\itemsep{0pt}
  \item the literal error rate: vector sentences whose translation has a
        literal word that the reference does not have (948 sentences);
  \item the meaning-loss rate: full-verb sentences whose translation
        misses the literal meaning.
\end{itemize}
We also report chrF++ \citep{popovic-2017-chrf}. We leave out \bn{পাঠানো}
(send), because its vector use keeps the sense of sending.

\paragraph{Statistics.}
All accuracies come with 95\% bootstrap confidence intervals over items.
Model comparisons use paired bootstrap tests.
\section{Results}
\label{sec:results}

\subsection{RQ1: Is the distinction encoded?}
\label{sec:rq1}

\begin{figure}[H]
  \centering
  \includegraphics[width=\columnwidth]{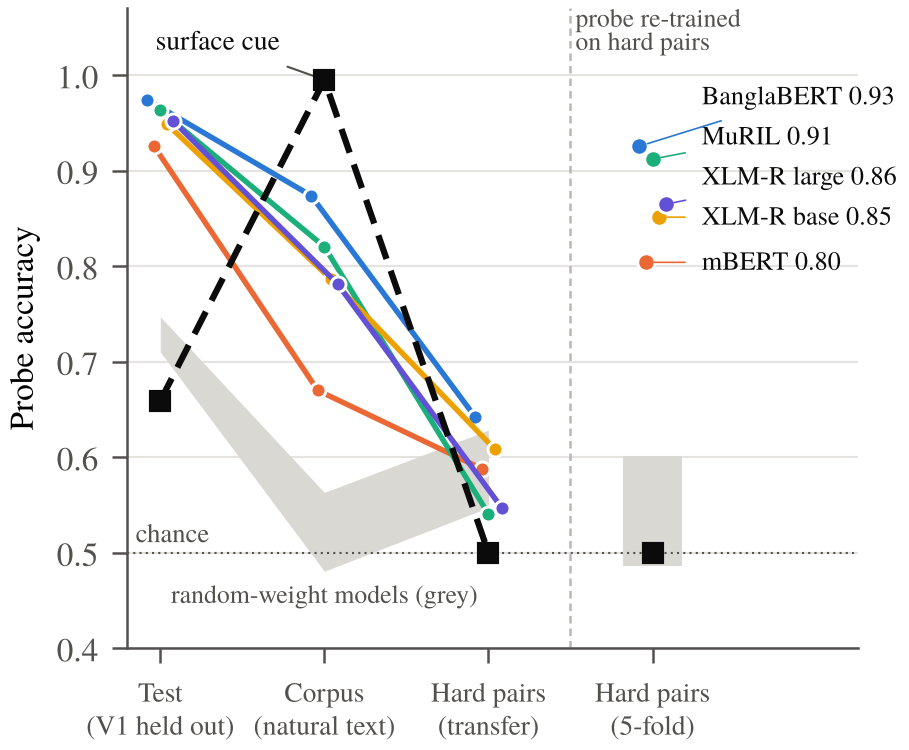}
  \caption{Probe accuracy on the test sets. Lines join results of the same
  probe, trained on ordinary pairs. The surface cue (dashed) is near-perfect on
  corpus pairs but at chance on hard pairs, and the transferred probe falls to
  the level of random-weight copies (grey). Probes re-trained within the hard
  pairs (5-fold, right) reach 0.80--0.93.}
  \label{fig:shortcut}
\end{figure}

\begin{table}[H]
\centering
\small
\setlength{\tabcolsep}{3pt}
\begin{tabular}{@{}lcccccc@{}}
\toprule
 & & & & \multicolumn{3}{c}{Hard pairs} \\
\cmidrule(l){5-7}
Model & $\ell$ & Test & Corpus & transfer & 5-fold & rand. \\
\midrule
BanglaBERT  & 4 & \textbf{.97} & \textbf{.87} & \textbf{.64} & \textbf{.93} & .51 \\
mBERT       & 9 & .93 & .67 & .59 & .80 & .51 \\
MuRIL       & 2 & .96 & .82 & .54 & .91 & .49 \\
XLM-R base  & 6 & .95 & .79 & .61 & .85 & .56 \\
XLM-R large & 9 & .95 & .78 & .55 & .86 & .60 \\
\midrule
Surface cue & -- & .66 & .995 & .50 & .50 & -- \\
Majority    & -- & .50 & .50 & .50 & .50 & -- \\
\bottomrule
\end{tabular}
\caption{Probe accuracy. $\ell$: layer chosen on dev. Transfer: probe
trained on ordinary pairs, at layer $\ell$; 5-fold: trained and tested
within the hard pairs; rand.: the same 5-fold probe on a random-weight
copy. The surface cue is correct on 205 of 206 corpus sentences (.995). All
pretrained--random differences in the 5-fold setting are significant
($p<0.001$).}
\label{tab:rq1}
\end{table}

\paragraph{High test accuracy alone is misleading.}
All five encoders separate vector from full-verb uses on the test set
(0.93--0.97; Table~\ref{tab:rq1}), and they reach 0.88--0.97 by layer~2
(Figure~\ref{fig:app-layers}). Two baselines show why this number cannot be
taken at face value (Figure~\ref{fig:shortcut}). Random-weight copies of the
same models reach 0.71--0.75, and the one-feature surface cue reaches 0.66 on
the test set and 0.995 on the corpus pairs. On real text, it is enough to ask
whether the word before V2 is a conjunctive participle.

\paragraph{Without the cue, the distinction is still there.}
On the hard pairs, the surface cue falls to 0.50. A probe trained on ordinary
pairs does not transfer to them: at the layer chosen on the development set it
scores 0.54--0.64, the level of random-weight copies (0.55--0.63;
Table~\ref{tab:app-rq1}). Probes trained and tested within the hard pairs
(5-fold), however, reach 0.80--0.93, while random-weight copies stay at
0.49--0.60 ($p<0.001$ for every model). BanglaBERT is best (0.93), followed by
MuRIL (0.91), XLM-R large (0.86), XLM-R base (0.85) and mBERT (0.80). The
contextual information is thus present, but it is not what a probe trained on
ordinary pairs relies on. Choosing the layer post hoc on the hard pairs
themselves, which is optimistic, raises transfer to 0.72--0.80 for four
encoders (BanglaBERT 0.80 at layer 9 of 12), but only to 0.61 for MuRIL
(Figure~\ref{fig:app-layers}). For BanglaBERT, mBERT and XLM-R base, the best
layers are in the middle or upper part of the network; MuRIL peaks at layer~1,
and XLM-R large ties between layer~1 and layers 19--20. Early layers thus track
the surface cue in most, but not all, encoders.
\subsection{RQ2: Is the meaning inferred?}
\label{sec:rq2}

\begin{figure*}[t]
  \centering
  \includegraphics[width=\textwidth]{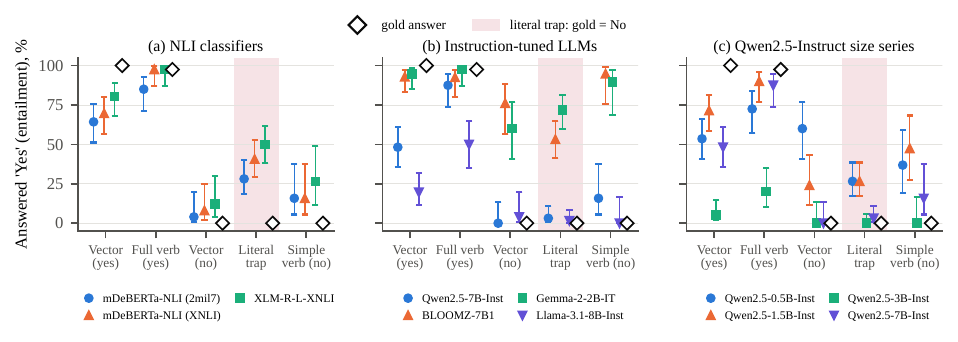}
  \caption{Share of \textit{Yes} answers per item category, with Wilson 95\%
  intervals. Hollow diamonds mark the gold answer. Shaded: literal traps, where
  the hypothesis states the V2's lexical meaning and the gold answer is
  \textit{No}. Several LLMs give nearly the same answer in every category.}
  \label{fig:rq2}
\end{figure*}

\begin{table}[t]
\centering
\small
\setlength{\tabcolsep}{3.5pt}
\begin{tabular}{@{}lcccc@{}}
\toprule
Model & Acc & AUC & Yes & Trap \\
\midrule
mDeBERTa-NLI (2mil7)  & .77 & .85 & .45 & .72 \\
mDeBERTa-NLI (XNLI)   & .77 & .84 & .53 & .59 \\
XLM-R-large-XNLI      & .75 & .86 & .61 & .50 \\
\midrule
Qwen2.5-0.5B-Inst     & .62 & .66 & .48 & .73 \\
Qwen2.5-1.5B-Inst     & .75 & .81 & .53 & .73 \\
Qwen2.5-3B-Inst       & .59 & .79 & .05 & 1.00 \\
Qwen2.5-7B-Inst       & \textbf{.81} & \textbf{.91} & .33 & .97 \\
BLOOMZ-7B1            & .61 & .80 & .78 & .47 \\
Gemma-2-2B-IT         & .59 & .74 & .83 & .28 \\
Llama-3.1-8B-Inst     & .68 & .88 & .16 & .98 \\
\midrule
always \textit{No}    & .53 & .50 & .00 & 1.00 \\
\bottomrule
\end{tabular}
\caption{Entailment results. Yes = share of \textit{Yes} answers (gold: 0.47);
Trap = accuracy on the 64 literal traps. LLM scores average an English and a
Bangla prompt.}
\label{tab:rq2}
\end{table}

\paragraph{Ranking is better than answering.}
Every model ranks entailed hypotheses above non-entailed ones better than
chance (AUC 0.66--0.91; Table~\ref{tab:rq2}). But accuracy only ranges from
0.59 to 0.81. The best model, Qwen2.5-7B-Inst, is not significantly more
accurate than ready-made NLI classifiers (0.81 vs.\ 0.77, $p=0.21$). Even the
basic inference behind our running example is hard. Nine of ten models deny
that \textit{no rice is left} after \bn{পুরো ভাতটা খেয়ে ফেলল} (ate up all the
rice). The only exception is Gemma-2-2B-IT, which says \textit{Yes} to 83\% of
all items.

\paragraph{Answer bias drives the errors.}
Several LLMs give the same answer to almost every item (Figure~\ref{fig:rq2}).
Llama-3.1-8B-Inst and Qwen2.5-3B-Inst answer \textit{No} to 84\% and 95\% of
items, even though their AUCs are 0.88 and 0.79. Gemma-2-2B-IT and BLOOMZ-7B1
answer \textit{Yes} to about 80\%. The \textit{Yes}-biased models accept the
literal meaning of the vector. Gemma says \textit{Yes} to 72\% of the literal
traps. For example, from \bn{শিশুটা হঠাৎ কেঁদে উঠল} (the child suddenly burst
into tears) it infers that the child climbed somewhere. The literal traps that
most models fall for involve \bn{মরা} read as die, \bn{পাঠানো} as send and
\bn{ছাড়া} as leave. Models also rarely tell a vector premise from its matched
control: pair consistency is at most 0.57.

\paragraph{The prompt language changes the answer.}
Changing only the instruction from English to Bangla flips several models.
Qwen2.5-0.5B-Inst goes from 10\% to 100\% \textit{Yes}, and Llama-3.1-8B-Inst
from 36\% to 2\%, while its AUC drops from 0.89 to 0.66
(Table~\ref{tab:app-rq2-prompt}). Only Qwen2.5-7B-Inst and BLOOMZ-7B1 stay
stable.
\subsection{RQ3: Are the combination rules known?}
\label{sec:rq3}
\begin{figure}[t]
  \centering
  \includegraphics[width=\columnwidth]{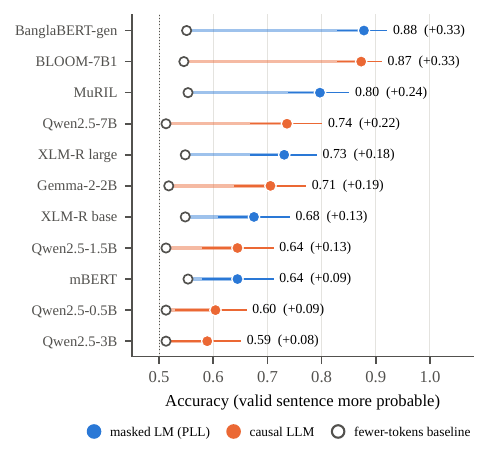}
  \caption{Selectional accuracy (filled) against each model's fewer-tokens
  baseline (hollow), with 95\% CIs. All models beat their baseline.}
  \label{fig:rq3}
\end{figure}

\paragraph{All models prefer the valid sentence.}
All eleven models score above chance and above their own fewer-tokens baseline
(Figure~\ref{fig:rq3}, Table~\ref{tab:rq3}), and per-token scores give a broadly similar ranking. The
preference therefore does not come from sentence length.

\begin{table}[t]
\centering
\small
\setlength{\tabcolsep}{3.5pt}
\begin{tabular}{@{}lcccc@{}}
\toprule
Model & Acc & Per tok. & Base & Benef. \\
\midrule
BanglaBERT-gen & \textbf{.88} & \textbf{.82} & .55 & .70 \\
mBERT          & .64 & .59 & .55 & .30 \\
MuRIL          & .80 & .77 & .55 & \textbf{.83} \\
XLM-R base     & .68 & .65 & .55 & .65 \\
XLM-R large    & .73 & .70 & .55 & .57 \\
\midrule
Qwen2.5-0.5B   & .60 & .55 & .51 & .39 \\
Qwen2.5-1.5B   & .64 & .58 & .51 & .35 \\
Qwen2.5-3B     & .59 & .56 & .51 & .43 \\
Qwen2.5-7B     & .74 & .69 & .51 & .48 \\
BLOOM-7B1      & .87 & .81 & .55 & .57 \\
Gemma-2-2B     & .71 & .58 & .52 & .43 \\
\bottomrule
\end{tabular}
\caption{Selectional accuracy (197 pairs). Per tok.: per-token log-probability;
Base: the model's fewer-tokens baseline; Benef.: the 23 benefactive pairs.}
\label{tab:rq3}
\end{table}

\paragraph{Bangla exposure matters more than size.}
The two best models are tied ($p=0.95$): the small Bangla-only BanglaBERT
generator (0.88) and BLOOM-7B1 (0.87), whose training data explicitly includes
Bangla. At the same size, BLOOM beats Qwen2.5-7B by 0.14 ($p<0.001$). Within the
Qwen series, accuracy is flat from 0.5B to 3B (0.60, 0.64, 0.59) and rises only
at 7B (0.74). Among encoders, the Bangla and Indic models (BanglaBERT 0.88,
MuRIL 0.80) lead mBERT (0.64) and XLM-R (0.68--0.73).

\paragraph{Local choices are easy; context is hard.}
When the valid and invalid sentences differ only in the vector itself
(\bn{দিলেন} vs.\ \bn{নিলেন}), models are at 0.79--1.00. When V1+V2 is the same
and only the context makes it ungrammatical, as with a durative adverbial on
completive \bn{ফেলা}, accuracy drops to 0.54--0.85. The weakest rule is
benefaction: most models are at or below chance (mBERT 0.30, Qwen 0.35--0.48,
Gemma 0.43). On the items where \bn{দেওয়া} is used with the agent as its own
beneficiary (\bn{নিজেকে}), every model scores 0.21--0.64.

\subsection{RQ4: Is the vector kept in translation?}
\label{sec:rq4}

\begin{figure*}[t]
  \centering
  \includegraphics[width=\textwidth]{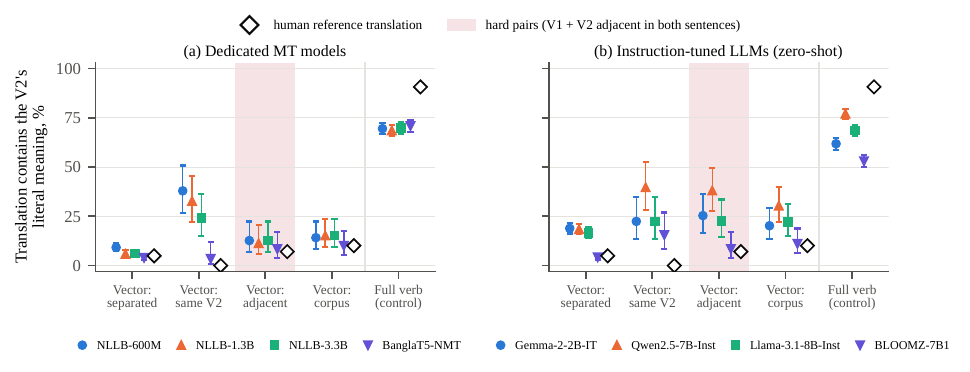}
  \caption{Share of English translations that contain the V2's literal meaning,
  by sentence type, with Wilson 95\% intervals. Hollow diamonds: human reference
  translations. For vector sentences, lower is better; for full-verb sentences
  (right), higher is better.}
  \label{fig:rq4}
\end{figure*}

\begin{table}[t]
\centering
\small
\setlength{\tabcolsep}{3.5pt}
\begin{tabular}{@{}lcccc@{}}
\toprule
 & Literal & Meaning & \multicolumn{2}{c}{chrF++} \\
\cmidrule(l){4-5}
System & error $\downarrow$ & lost $\downarrow$ & vec. & full \\
\midrule
NLLB-200 600M     &  9.6 & 27.1 & 45.9 & 50.9 \\
NLLB-200 1.3B     &  6.5 & 27.9 & 48.9 & 54.6 \\
NLLB-200 3.3B     &  6.5 & 26.4 & \textbf{50.3} & 55.7 \\
BanglaT5-NMT      & \textbf{2.8} & 25.6 & 49.9 & \textbf{59.7} \\
\midrule
Gemma-2-2B-IT     & 18.0 & 34.7 & 41.6 & 41.1 \\
Qwen2.5-7B-Inst   & 19.7 & \textbf{19.2} & 47.4 & 51.3 \\
Llama-3.1-8B-Inst & 15.4 & 27.6 & 49.5 & 52.5 \\
BLOOMZ-7B1        &  4.0 & 43.6 & 44.8 & 44.3 \\
\bottomrule
\end{tabular}
\caption{Translation results (\%). Literal error: vector sentences translated
with the V2's literal meaning. Meaning lost: full-verb sentences whose literal
meaning is missing. chrF++ against our reference translations.}
\label{tab:rq4}
\end{table}

\paragraph{LLMs translate vectors literally.}
Dedicated MT systems rarely add the V2's literal meaning to a vector sentence
(Table~\ref{tab:rq4}):
2.8\% for BanglaT5 and 6.5--9.6\% for NLLB-200. Instruction-tuned LLMs do so far
more often: 15.4\% for Llama-3.1-8B, 18.0\% for Gemma-2-2B and 19.7\% for
Qwen2.5-7B, the model that was best at entailment. Typical errors read the V2 as
a second event: Gemma renders \bn{রাগে সে চিঠিটা পুড়িয়ে ফেলল} `in anger she
burned up the letter' as \textit{He threw the letter in anger}, and Qwen
translates \bn{ভুল করে ফোনটা ভেঙে ফেলল} `accidentally broke the phone' as
\textit{broke the phone by throwing it}. BLOOMZ has a low literal rate (4.0\%),
but only because it often drops the verb's meaning altogether. It loses the
literal meaning of 43.6\% of full-verb sentences, the highest of all
systems.

\paragraph{The same verbs fail across tasks.}
Errors concentrate on a few vectors (Figure~\ref{fig:rq4}; per-verb results in
Appendix~\ref{app:results}). Futile \bn{মরা} is read as `die' by almost every
system, up to 55\% of the time for Qwen2.5-7B and 35--48\% for NLLB. Attitudinal
\bn{বসা} is read as `sit' by Qwen (49\%) and Llama (38\%). These are the same
vectors whose literal meaning models accepted in the entailment task
(\ref{sec:rq2}). The errors also occur in real text: on corpus sentences,
literal rates are 7--27\%.

\section{Discussion}
\label{sec:discussion}

\paragraph{Encoded, but not used.}
Our probes show that the vector/full-verb difference can be read from the
representations of all five encoders, even in pairs where no surface cue helps
(\S\ref{sec:rq1}). The behaviour tests show that prompted models rarely use
it. They miss inferences that vectors allow (\S\ref{sec:rq2}), they are
unreliable when only the context rules a vector out (\S\ref{sec:rq3}), and
they translate vectors as separate events (\S\ref{sec:rq4}). These two
findings come mostly from different models: we probed encoders, while the
clearest failures come from instruction-tuned LLMs, whose hidden states we did
not probe. So we do not claim that one LLM knows the difference and then
ignores it. We claim that the difference can be learned from Bangla text, yet
it is still lost when models are prompted. The closest case within one model
family points the same way. XLM-R large separates the two uses on hard pairs
(0.86), yet the XNLI classifier fine-tuned from it accepts half of the literal
traps. High probing accuracy should therefore not be taken as proof that a
system built on a model understands the construction.

\paragraph{The literal reading is systematic.}
The errors are not spread evenly across verbs. \bn{মরা} (die), \bn{পাঠানো}
(send) and \bn{ছাড়া} (leave) are the literal traps that most models accept in
the entailment task. \bn{মরা} and \bn{বসা} (sit) are also among the vectors
most often translated literally. So two very different tasks fail on some of
the same verbs and in the same direction, which points to a bias towards the
own meaning of V2. A similar bias towards the most common meaning has been
reported for Bangla words that are both names and common nouns
\citep{shuvo-islam-2026-name}. This suggests a wider pattern in how current
models handle Bangla words with a literal and a non-literal reading.

\paragraph{Bangla data often matters more than size.}
In three of the four tasks, the strongest systems are the ones built for
Bangla:
\begin{itemize}
  \setlength\itemsep{0pt}
  \item BanglaBERT leads in probing and is tied for first on selectional
        preferences;
  \item BanglaT5 makes the fewest literal translations;
  \item BLOOM, whose training data includes Bangla, beats Qwen2.5 of the same
        size by 14 points on selectional preferences.
\end{itemize}
The exception is entailment, where the best model is Qwen2.5-7B-Inst. Within
the Qwen2.5 series, size helps mainly at 7B.

\paragraph{Knowledge in probabilities may not carry over to instructions.}
BLOOM shows this, although the two tasks are different. As a base model it is
among the best at preferring correct combinations (0.87), but its
instruction-tuned version, BLOOMZ, is among the weakest at entailment (0.61,
with 78\% \textit{Yes} answers). It also avoids literal translations only by
dropping meaning. More generally, the instruction-tuned LLMs are sensitive to
answer bias and to the language of the prompt, in ways that a probe or a
probability comparison does not show. Testing Bangla LLMs only through prompts
may therefore underrate what they know. Testing models only through probes or
probabilities would overrate what they can do.

\paragraph{Implications.}
For evaluation, our results argue for test items without surface cues. On
ordinary pairs, a one-feature rule reaches 0.995 on real text, so benchmarks
without hard pairs can give models credit for knowledge they do not use. In
practice, dedicated MT systems are clearly safer than general LLMs for
translating Bangla compound verbs, and LLM translations of Bangla should be
checked for literal readings of vectors.

\section{Conclusion}
\label{sec:conclusion}

We introduced \textsc{BanglaCV-Probe}, a benchmark of Bangla compound verbs
that tests whether language models represent, understand, restrict and
translate vector verbs. All five encoders we probe clearly capture the
difference between vector and full-verb uses, but prompted models rarely use
it. They miss inferences that vectors allow, are unreliable when only the
context rules a vector out, and translate vectors literally, and some of the
same vectors fail across tasks. In three of four tasks, models built for
Bangla beat larger multilingual ones. Future work could probe the hidden
states of the LLMs themselves, extend the benchmark to Hindi, Urdu and other
languages with compound verbs, and test whether definitions or few-shot
examples close the gap between knowing and using. We release the benchmark and code.\footnote{\url{https://github.com/ashuvo25/Pole_OR_Vector}}
\section*{Limitations}

\paragraph{Mostly written data.}
We wrote most items ourselves. The 103 pairs from real text show the same
main patterns, but natural text may contain contexts our items miss. All
items use standard colloquial Bangla (\textit{chalit}); dialects and the
formal \textit{sadhu} style are almost absent.

\paragraph{Probing covers encoders only.}
We probe five encoders but not the LLMs whose behaviour we test. So we cannot
say whether a given LLM knows the difference and fails to use it, or does not
know it at all.

\paragraph{Labels for pragmatic inferences.}
Some entailment items, such as \textit{ate up the rice} $\Rightarrow$
\textit{no rice is left}, rest on a strong default inference rather than
strict entailment. Agreement over all items was high ($\kappa = 0.92$), but
other annotators could judge a few of these labels differently.

\paragraph{Evaluation choices.}
We tested the LLMs zero-shot with one English and one Bangla instruction,
greedy decoding and 4-bit quantisation. Other settings could change single
scores, although we expect the large gaps to hold. Probes show only that
information can be read from a model, not that the model uses it.

\paragraph{Literal-translation detector.}
RQ4 relies on English word lists for each vector's own meaning. They separate
the two uses well on our references (5\% vs.\ 91\%), but can miss paraphrases
and sometimes flag a correct idiom (e.g.\ \textit{take a shower} for
\bn{স্নান করে নাও}).

\paragraph{Confounds and small groups.}
In benefactive selectional items, the incorrect sentence often replaces a name
with the reflexive \bn{নিজেকে}, so word frequency may play a part. Some rule
types and per-verb subsets are small ($n\leq10$ for two rule types; 2--12 hard
pairs per verb), so we do not read much into them.
\section*{Ethics Statement}

The corpus-attested sentences come from public-domain literature (Rabindranath
Tagore, Sarat Chandra Chattopadhyay) and from publicly available news articles.
Authored sentences describe everyday situations and contain no personal
information. The annotation was carried out by the authors, who are native
speakers of Bangla. We will release the authored data under CC BY 4.0. For news
sentences, we will release only what their licences permit. All models were
used under their respective licences and only for evaluation. The experiments
needed a few GPU-hours on two T4 GPUs. Our findings point to a risk for users:
LLM-based translation of Bangla can silently change the meaning of a sentence,
for instance turning `burned up the letter' into `threw the letter'.

\bibliography{custom}

\appendix

\section* {Appendix}
\label{sec:appendix}

\raggedbottom   

\section{The 17 vector verbs}
\label{app:vectors}

\begin{table}[H]
\centering
\small
\setlength{\tabcolsep}{3pt}
\begin{adjustbox}{max width=\columnwidth}
\begin{tabular}{@{}lllp{0.43\columnwidth}@{}}
\toprule
ID & V2 & Lexical & Contribution as a vector \\
\midrule
V01 & \bn{ফেলা}     & throw      & completion; doing without hesitation \\
V02 & \bn{দেওয়া}    & give       & action done for someone else \\
V03 & \bn{নেওয়া}    & take       & action done for oneself \\
V04 & \bn{বসা}      & sit        & sudden, inappropriate action \\
V05 & \bn{ওঠা}      & rise       & sudden onset (inception) \\
V06 & \bn{যাওয়া}    & go         & change of state; continuation \\
V07 & \bn{পড়া}      & fall       & suddenness; resulting state \\
V08 & \bn{রাখা}     & keep       & done in advance (resultative) \\
V09 & \bn{আসা}      & come       & long duration up to now \\
V10 & \bn{আনা}      & bring      & approaching completion \\
V11 & \bn{তোলা}     & lift       & attainment after effort \\
V12 & \bn{চলা}      & walk, go on & continuous action \\
V13 & \bn{পাঠানো}   & send       & action through someone else \\
V14 & \bn{বেড়ানো}  & roam       & iterative, aimless action \\
V15 & \bn{মরা}      & die        & futility \\
V16 & \bn{দাঁড়ানো} & stand      & entering a state (inchoative) \\
V17 & \bn{ছাড়া}     & leave      & persistence until the goal \\
\bottomrule
\end{tabular}
\end{adjustbox}
\caption{Vector verbs in \textsc{BanglaCV-Probe}, following
\citep{paul2003composition, ghosh-2014-making}. ``Lexical'' is the meaning as a full verb.}
\label{tab:vectors}
\end{table}

\section{Dataset details}
\label{app:data}

\paragraph{Minimal-pair types.}
In separated pairs, the full verb has its own argument or location
between V1 and V2. In same-V2 pairs, V2 is a plain main verb. In
adjacent pairs, V1 directly precedes V2 in both sentences
(Table~\ref{tab:pairtypes}).

\begin{table}[H]
\centering
\small
\setlength{\tabcolsep}{3pt}
\begin{adjustbox}{max width=\columnwidth}
\begin{tabular}{@{}lrrrp{0.36\columnwidth}@{}}
\toprule
Type & Auth. & Corpus & Total & Full-verb example \\
\midrule
Separated & 809 & 10 & 819 & \bn{কলাটা খেয়ে খোসাটা রাস্তায় ফেলল} \\
Same V2   &  58 & 93 & 151 & \bn{গাছ থেকে একটা আম পড়ল} \\
Adjacent  &  74 &  0 &  74 & \bn{কাগজটা দলা পাকিয়ে ফেলেছে ঝুড়িতে} \\
\midrule
Total     & 941 & 103 & 1,044 & \\
\bottomrule
\end{tabular}
\end{adjustbox}
\caption{Minimal pairs by type and source (Auth.\ = authored, Corpus =
real text), with an example full-verb sentence for each type. The 2,088
sentences have 6.6 words on average and contain 552 distinct V1s and 932
distinct V1+V2 combinations.}
\label{tab:pairtypes}
\end{table}

\begin{figure*}[t]
  \centering
  \includegraphics[width=\textwidth]{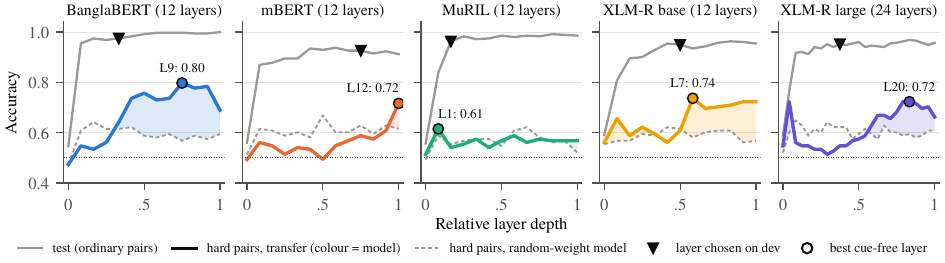}
  \caption{Layer-wise probe accuracy. Grey: test accuracy, which saturates by
  layer 2. Coloured: probe trained on ordinary pairs and applied to hard pairs;
  shading marks the gain over the random-weight copy (dashed).
  Triangles: layer chosen on dev; circles: best layer on hard pairs.}
  \label{fig:app-layers}
\end{figure*}
\paragraph{Removing shortcuts.}
Outside the adjacent pairs, the word before V2 is a conjunctive participle
(\textit{-e}) only in the vector sentence.
This one-feature cue scores 0.66 on
the test split and 0.995 on the corpus pairs. In an early version of the
adjacent pairs, V2 was also more often sentence-final in the vector sentence,
and this position cue scored 0.84. We rewrote 54 of the 74 pairs so that V2
has the same position in both sentences. Both cues now score 0.50, and
random-weight probes score 0.55--0.63.

\paragraph{Corpus pairs.}
We collected 260,063 sentences from public-domain fiction (Rabindranath Tagore,
Sarat Chandra Chattopadhyay) and from bdnews24 and Ei Samay news. We extracted
candidate V1+V2 sequences with regular expressions over NFC-normalised text,
which is needed for \bn{ড়} and \bn{য়}. Checking them by hand gave 103 pairs:
36 news--news, 28 literature--literature and 39 mixed.

\paragraph{Entailment items.}
Table~\ref{tab:nli} lists the categories. Twenty-one hypotheses occur after both
a vector and a control premise; we use these pairs for pair consistency.
\begin{table}[H]
\centering
\small
\setlength{\tabcolsep}{3pt}
\begin{adjustbox}{max width=\columnwidth}
\begin{tabular}{@{}lrl@{}}
\toprule
Category & $n$ & Gold \\
\midrule
Vector meaning (stated)        & 56 & entailment \\
Vector meaning (contradicted)  & 25 & not entailment \\
Literal trap                   & 64 & not entailment \\
Full-verb control              & 40 & entailment (39) \\
Simple-verb control            & 19 & not entailment \\
\midrule
Total                          & 204 & 95 / 109 \\
\bottomrule
\end{tabular}
\end{adjustbox}
\caption{Entailment items by category (11--13 per vector verb).}
\label{tab:nli}
\end{table}
\paragraph{Annotation.}
Three native speakers checked every item independently. They judged
naturalness and the intended reading for minimal pairs, the label for
entailment items, and the (in)validity for selectional pairs. Agreement was
Fleiss' $\kappa = 0.92$, and disagreements were resolved by discussion.

\paragraph{Selectional pairs.}
Of the 197 pairs, 6 come from \citep{paul2003composition}, 69 apply rules
stated there, and 122 are our own tests. There are 10--17 pairs per vector
verb (Table~\ref{tab:rules}).
 
\section{Additional results}
\label{app:results}

\subsection{RQ1: Probing}
\label{app:rq1}

\paragraph{Setup.}
For each sentence, we average the hidden states of the subword tokens of V2 at
every layer, including the embedding layer. We standardise these vectors and
train an L2-regularised logistic regression, with $C \in \{0.01, 0.1, 1, 10\}$.
The layer and $C$ are chosen on the development split. In the 5-fold
cross-validation on hard pairs, both are chosen inside each training fold, and
the two sentences of a pair are always in the same fold. Random-weight baselines
use the same architecture and tokenizer with freshly initialised weights.
Confidence intervals come from 2,000 bootstrap resamples over pairs.

\begin{table}[H]
\centering
\small
\setlength{\tabcolsep}{3pt}
\begin{adjustbox}{max width=\columnwidth}
\begin{tabular}{@{}lrp{0.44\columnwidth}@{}}
\toprule
Rule type & $n$ & Example of the constraint \\
\midrule
Duration / telicity        & 73 & completive \bn{ফেলা} rejects `for an hour' \\
Intentionality / futility  & 30 & persistive \bn{ছাড়া} rejects `by accident' \\
V1 verb class              & 28 & stative V1s reject certain vectors \\
Agentivity of V1           & 28 & some vectors need an agentive V1 \\
Benefactive direction      & 23 & \bn{দেওয়া} needs a beneficiary other than the agent \\
Arguments / animacy        & 10 & animacy or argument restrictions \\
\bn{হয়ে} frame             &  5 & \bn{হয়ে} + vector constructions \\
\midrule
Total                      & 197 & \\
\bottomrule
\end{tabular}
\end{adjustbox}
\caption{Selectional pairs by rule type. The two sentences of a pair differ in a
single word or phrase.}
\label{tab:rules}
\end{table}

\begin{figure}[H]
  \centering
  \includegraphics[width=\columnwidth]{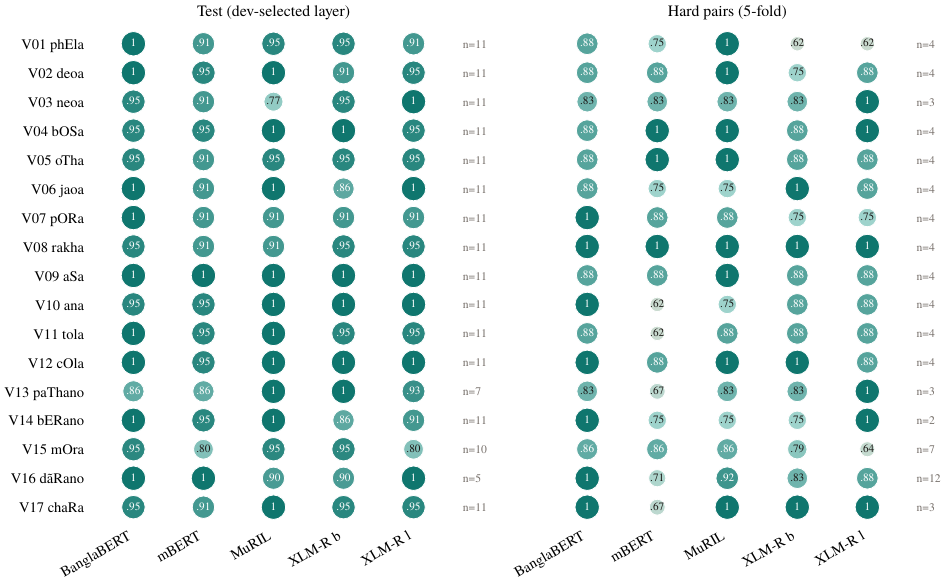}
  \caption{RQ1 probe accuracy per vector verb. $n$ = number of pairs. The hard
  pairs are few per verb, so per-verb differences there are indicative only.}
  \label{fig:app-rq1-vec}
\end{figure}

\subsection{RQ2: Entailment}
\label{app:rq2}

\paragraph{Prompts.}
Each item is wrapped in the model's chat template (BLOOMZ has none, so we append \textit{Answer:}). We use two instructions:

\begin{itemize}
  \setlength\itemsep{0pt}
  \item \textit{English:} ``Premise: \{p\} \textbackslash n Hypothesis: \{h\}
        \textbackslash n\textbackslash n If the premise is true, must the
        hypothesis also be true? Answer with Yes or No only.''
  \item \textit{Bangla:} ``\bn{বাক্য ১:} \{p\} \textbackslash n \bn{বাক্য ২:} \{h\}
        \textbackslash n\textbackslash n \bn{বাক্য ১ সত্য হলে বাক্য ২-ও কি অবশ্যই
        সত্য? শুধু} Yes \bn{অথবা} No \bn{লিখে উত্তর দিন।}''
\end{itemize}

\begin{table}[H]
\centering
\small
\setlength{\tabcolsep}{2.5pt}
\begin{adjustbox}{max width=\columnwidth}
\begin{tabular}{@{}lcccc@{}}
\toprule
Model & Test & Corpus & Hard (transfer) & Hard (5-fold) \\
\midrule
\multicolumn{5}{@{}l}{\textit{Pretrained}} \\
BanglaBERT  & .97 [.96,.99] & .87 [.83,.91] & .64 [.59,.70] & .93 [.89,.97] \\
mBERT       & .93 [.90,.95] & .67 [.62,.72] & .59 [.53,.65] & .80 [.75,.86] \\
MuRIL       & .96 [.94,.98] & .82 [.77,.87] & .54 [.50,.58] & .91 [.86,.95] \\
XLM-R base  & .95 [.93,.97] & .79 [.73,.83] & .61 [.57,.66] & .85 [.79,.91] \\
XLM-R large & .95 [.93,.97] & .78 [.73,.83] & .55 [.51,.58] & .86 [.81,.91] \\
\multicolumn{5}{@{}l}{\textit{Random weights}} \\
BanglaBERT  & .75 [.70,.79] & .56 [.50,.63] & .62 [.56,.68] & .51 [.45,.58] \\
mBERT       & .71 [.66,.76] & .48 [.42,.54] & .61 [.55,.68] & .51 [.44,.57] \\
MuRIL       & .72 [.68,.76] & .52 [.46,.58] & .55 [.48,.61] & .49 [.41,.57] \\
XLM-R base  & .72 [.67,.76] & .52 [.47,.57] & .57 [.51,.62] & .56 [.49,.64] \\
XLM-R large & .74 [.70,.78] & .53 [.47,.59] & .63 [.57,.69] & .60 [.53,.67] \\
\bottomrule
\end{tabular}
\end{adjustbox}
\caption{RQ1 probe accuracy with 95\% bootstrap confidence intervals. For the
5-fold evaluation, pretrained models beat their random copies with $p<0.001$
(paired bootstrap).}
\label{tab:app-rq1}
\end{table}

$P(\textit{Yes})$ sums the probabilities of the first tokens of
\textit{Yes/yes/YES} (with and without a leading space) and renormalises over
\textit{Yes} and \textit{No}. On average, 82--100\% of the next-token
probability falls on these answer tokens with the English prompt. With the
Bangla prompt it is 55--100\%, except for Qwen2.5-1.5B (16\%).

\begin{table}[H]
\centering
\small
\setlength{\tabcolsep}{2.5pt}
\begin{adjustbox}{max width=\columnwidth}
\begin{tabular}{@{}lcccccc@{}}
\toprule
 & \multicolumn{2}{c}{Acc} & \multicolumn{2}{c}{AUC} & \multicolumn{2}{c}{Yes-rate} \\
\cmidrule(lr){2-3}\cmidrule(lr){4-5}\cmidrule(l){6-7}
Model & en & bn & en & bn & en & bn \\
\midrule
Qwen2.5-0.5B-Inst & .63 & .47 & .64 & .58 & .10 & 1.00 \\
Qwen2.5-1.5B-Inst & .73 & .61 & .80 & .68 & .49 & .70 \\
Qwen2.5-3B-Inst   & .69 & .53 & .82 & .76 & .16 & .00 \\
Qwen2.5-7B-Inst   & .80 & .79 & .92 & .89 & .33 & .32 \\
BLOOMZ-7B1        & .63 & .62 & .79 & .79 & .76 & .78 \\
Gemma-2-2B-IT     & .55 & .64 & .80 & .71 & .91 & .66 \\
Llama-3.1-8B-Inst & .79 & .53 & .89 & .66 & .36 & .02 \\
\bottomrule
\end{tabular}
\end{adjustbox}
\caption{RQ2 prompt sensitivity: English vs.\ Bangla instruction.}
\label{tab:app-rq2-prompt}
\end{table}

\begin{table}[H]
\centering
\small
\setlength{\tabcolsep}{3pt}
\begin{adjustbox}{max width=\columnwidth}
\begin{tabular}{@{}lccc@{}}
\toprule
Model & Acc [95\% CI] & Macro-F1 & Pair cons. \\
\midrule
mDeBERTa-NLI (2mil7) & .77 [.71,.82] & .77 & .48 \\
mDeBERTa-NLI (XNLI)  & .77 [.71,.82] & .77 & \textbf{.57} \\
XLM-R-large-XNLI     & .75 [.69,.81] & .75 & .48 \\
Qwen2.5-0.5B-Inst    & .62 [.55,.69] & .62 & .19 \\
Qwen2.5-1.5B-Inst    & .75 [.69,.81] & .75 & .33 \\
Qwen2.5-3B-Inst      & .59 [.52,.65] & .47 & .14 \\
Qwen2.5-7B-Inst      & \textbf{.81} [.76,.86] & \textbf{.81} & .43 \\
BLOOMZ-7B1           & .61 [.54,.68] & .59 & .05 \\
Gemma-2-2B-IT        & .59 [.52,.66] & .55 & .00 \\
Llama-3.1-8B-Inst    & .68 [.61,.74] & .62 & .14 \\
\midrule
always \textit{No}           & .53 & .35 & -- \\
always \textit{Yes}          & .47 & .32 & -- \\
vector premise $\Rightarrow$ \textit{Yes} & .37 & .35 & -- \\
\bottomrule
\end{tabular}
\end{adjustbox}
\caption{RQ2 accuracy with CIs, macro-F1, and pair consistency (both items of
the 21 matched vector/control pairs correct).}
\label{tab:app-rq2-f1}
\end{table}

\begin{figure}[H]
  \centering
  \includegraphics[width=\columnwidth]{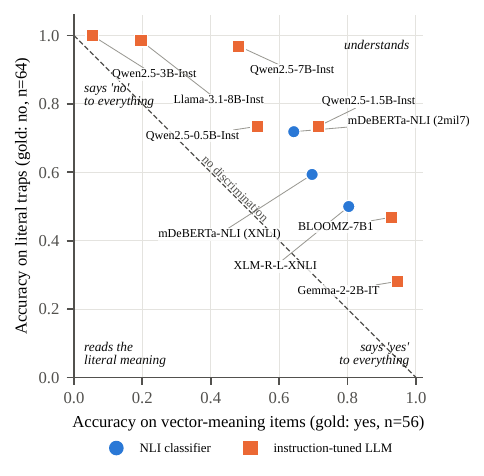}
  \caption{Accuracy on vector-meaning items (gold \textit{Yes}) vs.\ literal
  traps (gold \textit{No}). Points near the dashed line do not discriminate.}
  \label{fig:app-rq2-trap}
\end{figure}

\subsection{RQ3: Selectional preferences}
\label{app:rq3}

\begin{table*}[t]
\centering
\small
\setlength{\tabcolsep}{3.2pt}
\begin{adjustbox}{max width=\textwidth}
\begin{tabular}{@{}lrccccccccccc@{}}
\toprule
 & & \multicolumn{5}{c}{Masked LMs} & \multicolumn{6}{c}{Causal LMs} \\
\cmidrule(lr){3-7}\cmidrule(l){8-13}
 & $n$ & BnBERT-g & mBERT & MuRIL & XLM-R b & XLM-R l & Qw-0.5 & Qw-1.5 & Qw-3 & Qw-7 & BLOOM & Gemma \\
\midrule
\multicolumn{13}{@{}l}{\textit{By rule type}} \\
Duration / telicity       & 73 & .89 & .73 & .78 & .63 & .81 & .66 & .73 & .53 & .75 & \textbf{.93} & .75 \\
Intentionality / futility & 30 & \textbf{.83} & .57 & .60 & .63 & .53 & .63 & .60 & .57 & .67 & .73 & .57 \\
V1 verb class             & 28 & .96 & .61 & .86 & .68 & .71 & .61 & .61 & .75 & .82 & \textbf{1.00} & .75 \\
Agentivity of V1          & 28 & \textbf{.96} & .82 & .93 & .86 & .82 & .68 & .71 & .82 & .89 & \textbf{.96} & .82 \\
Benefactive direction     & 23 & .70 & .30 & \textbf{.83} & .65 & .57 & .39 & .35 & .43 & .48 & .57 & .43 \\
Arguments / animacy       & 10 & .80 & .60 & .80 & .50 & .80 & .60 & .60 & .50 & .60 & \textbf{.90} & .80 \\
\bn{হয়ে} frame            &  5 & 1.00 & .80 & 1.00 & 1.00 & 1.00 & .20 & 1.00 & .20 & 1.00 & 1.00 & 1.00 \\
\midrule
\multicolumn{13}{@{}l}{\textit{By what differs between the valid and the invalid sentence}} \\
V2 changed                & 19 & \textbf{1.00} & .84 & .89 & .89 & .79 & .95 & .89 & .84 & .89 & \textbf{1.00} & .89 \\
V1 changed                & 32 & .94 & .66 & .84 & .81 & .81 & .53 & .53 & .66 & .91 & \textbf{.97} & .81 \\
Context changed           & 146 & \textbf{.85} & .62 & .77 & .62 & .71 & .58 & .64 & .54 & .68 & .84 & .66 \\
\bottomrule
\end{tabular}
\end{adjustbox}
\caption{RQ3 accuracy by rule type and by contrast type. The contrast type is
assigned automatically from the words that differ between the two sentences.
Cells with $n<10$ are not interpreted.}
\label{tab:app-rq3}
\end{table*}

\paragraph{Scoring.}
Causal models score each sentence after a beginning-of-sequence token (or the
end-of-sequence token if there is none). Masked models use PLL-word-l2r
\citep{kauf-ivanova-2023-better}: each subword is masked together with the
following subwords of the same word. Ties count as half correct.

\begin{table}[H]
\centering
\small
\begin{adjustbox}{max width=\columnwidth}
\begin{tabular}{@{}lrr@{}}
\toprule
Comparison & $\Delta$ & $p$ \\
\midrule
BanglaBERT-gen vs.\ BLOOM-7B1 & +0.005 & 0.95 \\
BLOOM-7B1 vs.\ Qwen2.5-7B     & +0.137 & $<$0.001 \\
Qwen2.5-7B vs.\ Qwen2.5-3B    & +0.147 & $<$0.001 \\
Qwen2.5-7B vs.\ Qwen2.5-0.5B  & +0.132 & $<$0.001 \\
BanglaBERT-gen vs.\ MuRIL     & +0.081 & 0.007 \\
MuRIL vs.\ XLM-R large        & +0.066 & 0.057 \\
XLM-R large vs.\ XLM-R base   & +0.056 & 0.12 \\
\bottomrule
\end{tabular}
\end{adjustbox}
\caption{RQ3 paired bootstrap comparisons (5,000 resamples over pairs).}
\label{tab:app-rq3-tests}
\end{table}

\begin{figure}[H]
  \centering
  \includegraphics[width=\columnwidth]{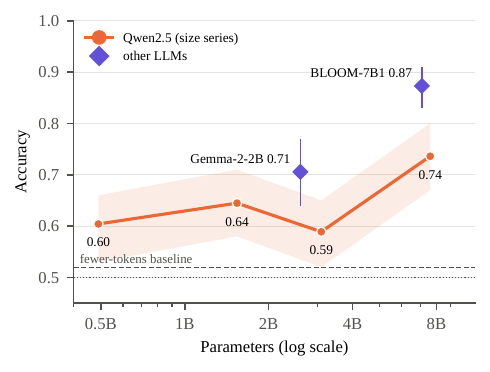}
  \caption{RQ3 accuracy of causal models by size. The Qwen2.5 series is flat up
  to 3B; BLOOM-7B1 clearly exceeds Qwen2.5-7B.}
  \label{fig:app-rq3-scaling}
\end{figure}

\subsection{RQ4: Translation}
\label{app:rq4}

\paragraph{Decoding.}
The MT systems use beam search with 4 beams and up to 96 new tokens. NLLB 
\begin{figure}[H]
  \centering
  \includegraphics[width=\columnwidth]{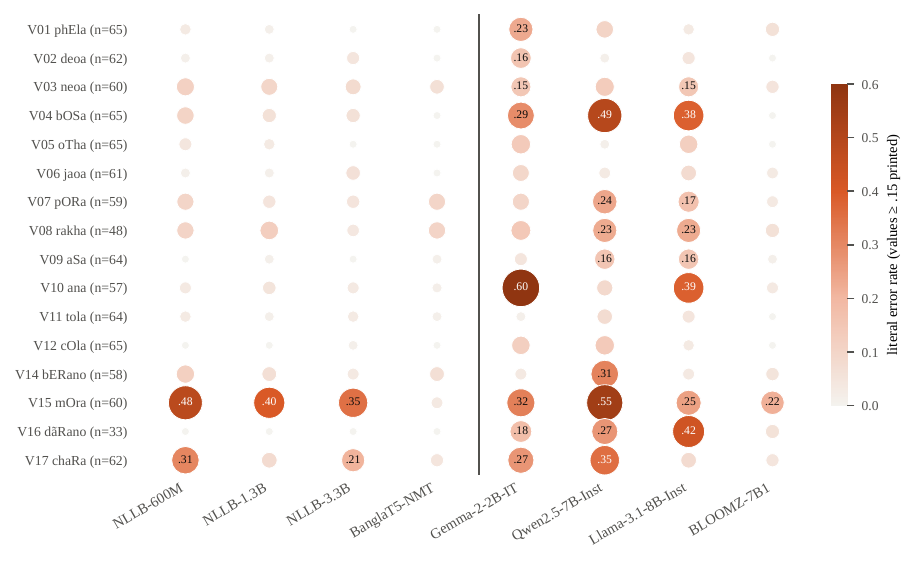}
  \caption{RQ4 literal error rate per vector verb (values $\geq$ .15 printed).
  \bn{মরা} `die' is read literally by almost every system.}
  \label{fig:app-rq4-vec}
\end{figure}

\begin{figure}[H]
  \centering
  \includegraphics[width=\columnwidth]{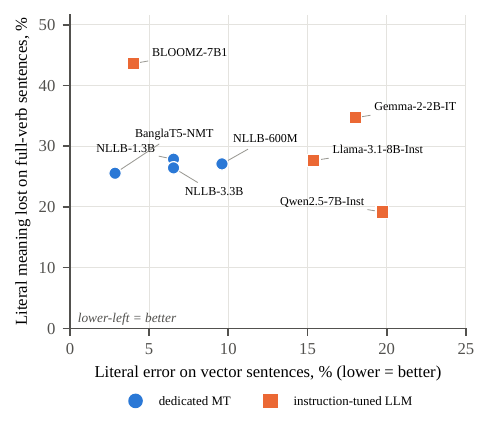}
  \caption{RQ4 trade-off between literal errors on vector sentences and lost
  meaning on full-verb sentences. BLOOMZ avoids literal errors by dropping
  meaning.}
  \label{fig:app-rq4-trade}
\end{figure}
uses \texttt{ben\_Beng}$\rightarrow$\texttt{eng\_Latn}. The LLMs use greedy decoding (up to 80
tokens) with the instruction ``Translate the following Bangla sentence into
English. Reply with the English translation only.'' We keep the first
non-empty line and strip prefixes such as \textit{Translation:}.
\paragraph{Literal-meaning detector.}
For each V2 we list the English forms of its lexical meaning
(Table~\ref{tab:app-lit}) and match them as whole words. \bn{পাঠানো} is
excluded, because its vector use (\bn{ডেকে পাঠাল} `sent for') keeps the sense of
sending.

\begin{table}[H]
\centering
\small
\setlength{\tabcolsep}{3pt}
\begin{adjustbox}{max width=\columnwidth}
\begin{tabular}{@{}lp{0.68\columnwidth}@{}}
\toprule
V2 & English forms counted as literal \\
\midrule
\bn{ফেলা} & throw, drop, toss, cast, discard, dump \\
\bn{দেওয়া} & give, hand \\
\bn{নেওয়া} & take \\
\bn{বসা} & sit, seated \\
\bn{ওঠা} & rise, climb, get up, go up, get into/on, board \\
\bn{যাওয়া} & go \\
\bn{পড়া} & fall \\
\bn{রাখা} & keep, put, place, store \\
\bn{আসা} & come \\
\bn{আনা} & bring, fetch \\
\bn{তোলা} & lift, raise, pick up, hoist, load, put, withdraw \\
\bn{চলা} & walk, move, head, set off, proceed, on one's way \\
\bn{বেড়ানো} & stroll, wander, roam, tour, travel, visit, walk \\
\bn{মরা} & die, dead, death, killed \\
\bn{দাঁড়ানো} & stand \\
\bn{ছাড়া} & leave, release, let go, quit, abandon, set free \\
\bottomrule
\end{tabular}
\end{adjustbox}
\caption{Literal-meaning word lists (all inflected forms included). On the
reference translations, they match 5\% of vector and 91\% of full-verb
sentences.}
\label{tab:app-lit}
\end{table}

\begin{table}[H]
\centering
\small
\setlength{\tabcolsep}{4pt}
\begin{tabular}{@{}lcccc@{}}
\toprule
 & \multicolumn{3}{c}{Authored} & Corpus \\
\cmidrule(lr){2-4}\cmidrule(l){5-5}
System & separated & same V2 & adjacent & (real text) \\
 & {\footnotesize $n{=}735$} & {\footnotesize $n{=}58$} & {\footnotesize $n{=}66$} & {\footnotesize $n{=}89$} \\
\midrule
\multicolumn{5}{@{}l}{\textit{MT systems}} \\
NLLB-0.6B    & 7 & 38 & 9 & 10 \\
NLLB-1.3B    & 4 & 33 & 8 & 10 \\
NLLB-3.3B    & 4 & 24 & 9 & 11 \\
BanglaT5     & \textbf{2} & \textbf{3} & \textbf{3} & \textbf{7} \\
\midrule
\multicolumn{5}{@{}l}{\textit{Instruction-tuned LLMs}} \\
Gemma-2-2B   & 18 & 22 & 21 & 16 \\
Qwen2.5-7B   & 16 & 40 & 36 & 27 \\
Llama-3.1-8B & 14 & 22 & 18 & 20 \\
BLOOMZ       & \textbf{2} & 16 & 6 & 9 \\
\bottomrule
\end{tabular}
\caption{RQ4 literal error rate (\%) by subset: vector sentences whose
reference contains no literal word, but whose translation does.}
\label{tab:app-rq4-subset}
\end{table}

\section{Models}
\label{app:models}

We use the following public checkpoints from Hugging Face.

\paragraph{Encoders.} BanglaBERT and its generator
\citep{bhattacharjee-etal-2022-banglabert}, mBERT
\citep{devlin-etal-2019-bert}, MuRIL \citep{khanuja2021muril}, and XLM-R base
and large \citep{conneau-etal-2020-unsupervised}.

\paragraph{NLI classifiers.} Two mDeBERTa-v3 models fine-tuned for NLI
\citep{he2023debertav3} and XLM-R large fine-tuned on XNLI
\citep{conneau-etal-2020-unsupervised}.

\paragraph{Decoder LLMs.} Qwen2.5 base and Instruct at 0.5B, 1.5B, 3B and 7B
\citep{qwen2025qwen25}; BLOOM-7B1 \citep{workshop2022bloom} and its
instruction-tuned version BLOOMZ-7B1 \citep{muennighoff-etal-2023-crosslingual};
Gemma-2-2B and Gemma-2-2B-IT \citep{gemmateam2024gemma2}; and
Llama-3.1-8B-Instruct \citep{grattafiori2024llama3}.

\paragraph{MT systems.} NLLB-200 at 600M (distilled), 1.3B (distilled) and
3.3B \citep{nllb2022}, and BanglaT5 fine-tuned for Bangla--English translation
\citep{bhattacharjee-etal-2023-banglanlg}.

\paragraph{Precision.} Models above 3B run in 4-bit NF4 precision
\citep{dettmers2023qlora}.

\end{document}